\documentclass{article}
\usepackage{proceed2e}

\usepackage{mathptmx}
\usepackage{latexsym} 

\usepackage{amsmath}
\usepackage{graphicx}
\usepackage[square]{natbib}
\usepackage{url}
\usepackage[algoruled,vlined,noend]{algorithm2e}

\newtheorem{exampl}{Example}
\newenvironment{example}{\begin{exampl}\em}{\end{exampl}}
\newcommand{\tuple}[1]{\langle #1\rangle}
\usepackage{mymathfont}
\usepackage{color}

\title{Towards Completely Lifted Search-based Probabilistic Inference}
\author{David Poole, Fahiem Bacchus, and Jacek Kisynski\\
\url{http://www.cs.ubc.ca/~poole/}\\
\url{http://www.cs.toronto.edu/~fbacchus/}\\
\url{http://www.cs.ubc.ca/~kisynski/}
}

\begin{document}

\maketitle

\begin{abstract}
  The promise of lifted probabilistic inference is to carry out
  probabilistic inference in a relational probabilistic model without
  needing to reason about each individual separately (grounding out
  the representation) by treating the undistinguished individuals as a
  block. Current exact methods still need to ground out in some cases,
  typically because the representation of the intermediate results is
  not closed under the lifted operations. We set out to answer the
  question as to whether there is some fundamental reason why lifted
  algorithms would need to ground out undifferentiated individuals.
  We have two main results: (1) We completely characterize the cases
  where grounding is polynomial in a population size, and show how we
  can do lifted inference in time polynomial in the logarithm of the
  population size for these cases. (2) For the case of no-argument and
  single-argument parametrized random variables where the grounding
  is not polynomial in a population size, we present lifted
  inference which is polynomial in the population size whereas
  grounding is exponential. Neither of these cases requires reasoning
  separately about the individuals that are not explicitly mentioned.
\end{abstract}

\section{Introduction}
The problem of lifted probabilistic inference in its general form was
first explicitly
proposed by \citet{reason:Poole03a}, who formulated the problem in
terms of parametrized random variables, introduced the use of splitting to
complement unification, the parfactor representation of intermediate
results, and an algorithm for multiplying factors in a lifted manner.
de Salvo Braz et al.\ [\citeyear{reason:BraAmiRot05a,de-Salvo-Braz:2007aa}] invented
counting elimination for some cases where grounding would create a factor
with size exponential in the number of individuals, but lifted
inference can be done by counting the number of individuals with each
assignment of values.
\citet{Milch:2008aa} proposed counting formulae as a
representation of the intermediate result of counting, which allowed
for more cases where counting was applicable. However, this
body of research has not fulfilled the promise of lifted inference, as the
algorithms still need to ground in some cases.  The main problem is that
the proposals are based on variable elimination
\citep{reason:ZhaPoo94a}. This is a dynamic programming approach which 
requires a representation of the intermediate results, and the current 
representations for such results are not closed under all of the operations used for
inference. We sought to investigate whether there were fundamental
reasons why we need to ground in some cases.

An alternative to variable elimination is to used search-based methods
based on conditioning
such as recursive conditioning \citep{reason:Darwiche2001a} and other
methods \citep[see e.g.,][]{Bacchus:2009aa}. The advantage
of these methods is that conditioning simplifies the representations,
rather than complicating them.
The use of lifted search-based inference was proposed by
\citet{gogate10exploiting}, however to be both correct and able to do
inference without grounding requires more attention to detail than given in
that paper. This paper answer different questions than \citet{Jha:2010aa}.

Note that this paper is about exact inference. Lifted algorithms based
on belief propagation (e.g. by \citet{Singla:2008aa} and \citet{kersting09uai}) explicitly
ignore the interdependence amongst the instances that exact inference
needs to take into account.

In deriving an algorithm that never needs to ground, it is often the
examples that demonstrate why simpler methods do not work that are most insightful. We
have thus chosen to write this paper by presenting examples that
exemplify the cases that need to be considered. 

\section{Background}
\begin{figure*}
\begin{center}
\includegraphics[scale=0.7]{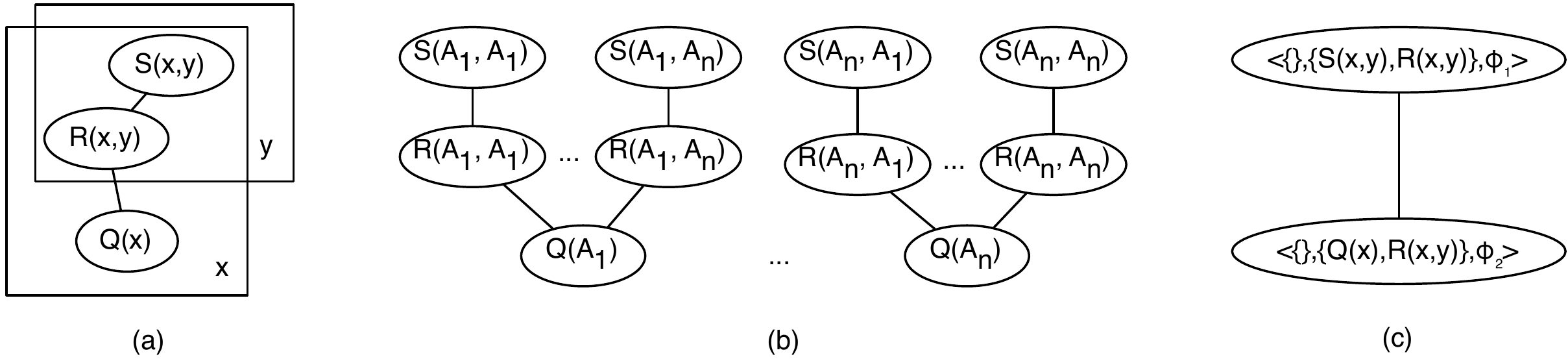}
\end{center}
\caption{(a) a parametrized graphical model, (b) its grounding and (c)
  its parfactor graph
}
\label{ExampleRQ}
\end{figure*}

The problem of lifted inference arises in relational probabilistic
models where there are probability distributions over random variables
that represent relations which depend on
individuals. \citet{reason:Poole03a} gives an example where the
probability that a person fitting a description committed a crime depends on the population
size, as this determines how many other people fit the
description. We don't want to reason about the other
individuals separately.
Rather, we would like to reason about them as a block considering only the number of such individuals.

A {\bf population} is a set of {\bf individuals}. A population
corresponds to a domain in logic. The {\bf population size} is  the
cardinality of the population which can be a finite number.\footnote{Infinite population sizes turn out to be simpler
  cases as $p^\infty=0$ for any $p<1$. So the infinite
  case allows for more pruning.} For the examples below, where there is a single population, we write the population as $A_1\dots A_n$, where $n$ is
the population size.

A \textbf{parameter}, which corresponds to a logical variable, is written in lower
case. Parameters are typed with a population; if $x$ is a parameter of
type $\tau$,
$pop(x)$ is the population associated with $x$ and $|x|=|\tau|=|pop(x)|$. We
assume that the populations are disjoint (and so the types are
mutually exclusive). 
Constants are written starting with an upper case letter. 

A \textbf{parametrized
random variable (PRV)} is of the form $F(t_1,\dots,t_k)$ where $F$ is a
k-ary functor (a function symbol or a predicate) and each $t_i$ is a
parameter or a constant. Each functor has a range, which is
$\{True,False\}$ for predicate symbols. A parametrized
random variable represents a set of random variables, one for each
assignment of an individual to a parameter. The range of the functor
becomes the domain of the random
variables.

A \textbf{substitution} is of the form $\{x_1/t_1,\dots,x_k/t_k\}$ where $x_i$
are distinct parameters and $t_i$ are constants or
parameters, such that $x_i$ and $t_i$ are of the same type. Given a PRV $r$ and a substitution $\theta=\{x_1/t_1,\dots,x_k/t_k\}$, the
application of $\theta$ on $r$, written
$r\theta$ is the PRV with each $x_i$ replaced by $t_i$. A substitution
$\theta = \{x_1/t_1,\dots,x_k/t_k\}$ \textbf{grounds} parameters
$x_1\dots x_k$ if $t_1\dots t_k$ are constants.
A
\textbf{grounding substitution} of $r$ is a substitution
that grounds all of the
parameters of $r$.

Probabilistic inference relies on knowing whether two random variables
are the same. With parametrized random variables, we unify them to
make them identical, but instead of just applying a substitution (as
in regular theorem proving), \citet{reason:Poole03a} proposed to split parametrized random
variables, forming the unifier and residual PRVs. 
\begin{example}
Applying substitution $\{x/A,y/B\}$ to PRV $Foo(x,y,z)$  results in PRV that is the direct application, $Foo(A,B,z)$, and
two ``residual'' PRVs, $Foo(x,y,z)$ with the constraint
$x\neq A$ and $Foo(A,y,z)$ with the constraint $y\neq B$; these three parametrized
random variables, with their associated constraints, represent the
same set of random variables as $Foo(x,y,z)$. 
\end{example}

A parametrized graphical model (Bayesian network or Markov network) is
a network with
parametrized random variables as nodes, and the instances of these share potentials. We need to be explicit
about which instances share potentials.

\begin{example}The parametrized graphical model of Figure
  \ref{ExampleRQ} (a), is shown using parametrized
random variables and plates (where the parameters
correspond to plates). The plate representation represents $n$
instances of $Q(x)$ and $n^2$ instances of $R(x,y)$ in the
grounding, shown in Figure \ref{ExampleRQ} (b). 
\end{example}

We assume the input to our algorithm is in the form of parfactors.  A {\bf parametric factor} or \textbf{parfactor}
\citep{reason:Poole03a} is a triple  $\tuple{C,V,\phi}$ where $C$ is a
set of inequality constraints on parameters, $V$ is a set
of parametrized random variables and $\phi$ is a factor, which is a function
from assignments of values to $V$ to the non-negative reals. $\phi$ is used as
the potential for all instances of the parfactor that are consistent
with the constraints.\footnote{This is known
  as parameter sharing, but where the parameters are the parameters of
the graphical model, not the individuals. Unfortunately, the logical
and probabilistic literature often uses the same terminology for
different things. Here we follow the traditions as much as seems
sensible, and apologize for any confusions. In particular ``$=$'' is used between a (parametrized) random variable and its value, whereas ``$\neq$'' and
``$/$'' are used for parameters (logical variables).} \citet{Milch:2008aa} also
explicitly include a set of parameters in their parfactors, but
we do not. A parfactor means its grounding; the set of factors on
$V\theta$ (all with table $\phi$) for each grounding
substitution $\theta$ of $V$ that obeys the constraints in $C$.

\subsection{Lifted Inference}
Lifted variable elimination, such as in C-FOVE \citep{Milch:2008aa},
allows for inference to work at the lifted level (doing unification
and splitting at runtime or as a preprocessing step) like normal variable elimination, until we 
remove a PRV that contains the only instance of a free
parameter or is linked to a PRV with a different set of parameters. At this stage, we need to take into account that the PRV
represents a set of random variables.

\begin{example}
Suppose we have a factor on $S(x,y)$, $R(x,y)$ and $Q(x)$, as in
Figure \ref{ExampleRQ} (a). We can sum out all instances of $S(x,y)$,
as all of the factors are the same, and get a new factor on $R(x,y)$;
this does $n^2$ (identical) operations on the grounding in a single step. If we
then remove $R$, we are multiplying a set of identical factors, and so can take their value to the power of the population
size \citep{reason:Poole03a}. 

Suppose, instead, we were to first sum out $Q(x)$. In the grounding, for each
individual $A_i$, the random variables $R(A_i,A_1)\dots R(A_i,A_n)$ are interdependent and so eliminating $Q$ results in a
factor on $R(A_i,A_1)\dots R(A_i,A_n)$. The size of this factor is
exponential in $n$. 
\end{example}

\Citet{reason:BraAmiRot05a} realized that the
identity of the individuals is not important; only the number of
individuals having each value of $R$. They introduced counting to solve
cases such as removing $Q(x)$ first in polynomial time rather than the exponential time (and
space) used for the
ground case. They, however, do the counting and summing
in one step, which limits its applicability. \citet{Milch:2008aa} defined counting formulae that give
a representation for the resulting lifted formula and can then be
combined with other factors. This expanded the applicability of lifted
inference, but it still requires grounding in some cases.
\subsection{Search-based probabilistic inference}
An alternative to lifting variable elimination is to lift a search-based method.
\begin{algorithm}
\DontPrintSemicolon
\SetAlgoNoEnd
\SetKwInput{Input}{input}
\SetKwInput{Output}{output}
\SetKwFunction{proc}{def}
\SetKwBlock{Begin}{}{}
\Input{$Con$: a  set of $variable=value$ assignments\;
$Fs$: set of factors
}
\Output{a number representing $\sum_{\overline{x}} \prod_{F \in Fs}
  F(\overline{x}|Con)$}
\If(\{\emph{Case 0}\})
{$vars(Con) \not\subseteq vars(Fs)$}{
\Return{$rc(\{(x=V)\in Con: x\in vars(Fs)\},Fs)$}
}
\If(\{\emph{Case 1}\})
{$\exists v$ such that $\tuple{\tuple{Con,Fs},v}\in cache$}{
\Return{$v$}}
\ElseIf(\{\emph{Case 2}\}){$\exists f\in Fs: vars(f)\subseteq vars(Con)$}{
     $F_0 \gets \{f \in Fs: vars(f)\subseteq vars(Con)\}$\;
     \Return{ $\left(\prod_{f \in F_0} eval(f,Con)\right) \times rc(Con,Fs\setminus F_0)$}\;
}
\ElseIf(\{\emph{Case 3}\}){factor graph $\tuple{Con,Fs}$ is disconnected}{
     $\Return \prod_{\mbox{connected component }cc}rc(Con,cc)$\;}
\Else(\{\emph{Case 4}\}){
select variable $x \in vars(Fs)\setminus vars(Con)$\;
$sum \gets 0$\;
\For{each value $v$ of $x$}{
$sum \gets sum+ rc(\{x=v\}\cup Con,Fs)$\;
}
$cache \gets cache \cup \{\tuple{\tuple{Con,Fs},sum}\}$\;
\Return{$sum$}\;
}
\caption{Recursive Conditioning: $rc(Con,Fs)$}
\label{RC-algo}
\end{algorithm}
The classic search-based algorithm is
recursive conditioning \citep{reason:Darwiche2001a}, a version of
which is presented in Algorithm \ref{RC-algo}.\footnote{Typically recursive conditioning requires computing a decomposition tree (D-Tree). Here we follow more the approach of \citep{Bacchus:2009aa} where we dynamically examine the problem for disconnected components as the search proceeds.} This
algorithm is presented in this non-traditional way, to emphasize the
cases that need to be implemented for lifting. In particular,  decoupling
branching and the evaluation of factors is useful for developing
its lifted counterpart. The correctness does not depend on the
order of the cases (although efficiency does).  In this algorithm $Con$ is
a context, a set of $variable=value$ assignments, and $Fs$ is
a set of input factors (this algorithm never creates or modifies factors; it only evaluate them when all of their variables are assigned). We separate the
context from the factors; typically these are combined to give
what could be called partially-assigned factors.\footnote{For the
  lifted case, projecting the context onto the factors loses
  information that is needed; see Footnote \ref{lifted-con-footnote}. It also makes it conceptually clearer
  that the factors share the same context.}

In case 0, if there are variables that appear in $Con$ that
do not appear in $Fs$, these are removed from $Con$. This is called
``forgetting'' in the description below; we forget the context that is
not relevant for the rest of the factors. $vars(S)$ is the set of variables that appear in $S$.

In case 1, the cache contains a set of previously
computed values. If a value has already been computed it can be recalled. Initially the cache is
$\{\left<\tuple{\{\},\{\}},1\right>\}$.

For case 2, if all of the variables that appear in a factor
$f$ are assigned in $Con$,  $eval(F,Con)$ returns the number that is the 
value of $F$ for the assignment $Con$.
These numbers are multiplied.

For case 3,  a
\textbf{factor graph}\footnote{This is related to a factor graph of
  \citet{Frey:2003aa} but we don't explicitly model the variable
  nodes.} on $\tuple{Con,Fs}$
is a graph where the nodes are factors in $Fs$, and there
is an arc between factors that share a random
variable that isn't assigned in $Con$. The connected components refer to the nodes that are
connected in this graph. The connected components can be solved
separately, and their return values multiplied.

Case 4 \textbf{branches} on a variable $x$ that isn't
assigned. 
The efficiency, but not the correctness, of the algorithm
depends on which variable $x$ is selected to be branched on.

To compute $P(x|Obs)$,  for each value $v$ of
$x$, call $rc(\{x=v\}\cup Obs,Fs)$ where $Fs$ is the set of all
factors of the model, and normalize the results.

An aspect that is important for lifted inference is that
when values are assigned, the factors are simplified as they are now
functions of fewer variables. This should be contrasted to variable
elimination that constructs more complicated factors.
\section{Search-based Lifted Inference}
In this section, we develop a lifted search-based algorithm. We
show its correctness with respect to a parallel ground algorithm that
uses the same order for splitting. Note that, because the lifted
algorithm removes multiple variables at once, this restricts
the order the variables are split in the ground algorithm. A legal
ordering for the lifted inference branches on all instances of a PRV
at once, whereas the corresponding ground algorithm branches on all
instances sequentially.
We
show how the complexity (as a function of the population size) of the
lifted algorithm is reduced compared to the corresponding ground
algorithm. Because we want our algorithm to be correct for all
legal branching orderings, we ignore the selection of which
variable to branch on; this can be optimized for efficiency.

We assume that we can count the number of solutions to a CSP with
inequality constraints in time that is at most logarithmic in the
domain size of the variables (e.g., by adapting the \#VE algorithm of
\citet{Dechter:2003aa} to not enumerate the undistinguished variables).

\subsection{Intermediate Representations}\label{IntRepSec}
The lifted analogy of a context in Algorithm \ref{RC-algo} is a
counting context which represents counts of
assignments to parametrized random variables.

A \textbf{counting context} on $V$ is a pair $\tuple{V,\chi}$, where $V$
is a set of PRVs (all taking a single argument of the same type), all parametrized by the same parameter, and $\chi$
is a table mapping assignments of PRVs in $V$ into non-negative integers. 
A counting context represents a context in the grounding. For each
individual of the type, the table $\chi$ specifies how many of the individuals
take on that tuple of values. We can also treat a counting context in
terms of $\chi$ as a set of
pairs of the assignment of values to $V$
and the corresponding count.

\begin{example}
A counting context for
$V=\{R(x),T(x)\}$, and
\begin{align*}
\chi=&\left\{\begin{array}{cc|c}
R(x)& T(x) & Value\\\hline
True & True & 20\\
True & False & 40\\
False & True & 10\\
False & False & 20
\end{array}\right.
\end{align*}
represents the assignment of values to $R$ and $T$ for $90$ ($20+40+10+20$) individuals. 20 of these have both $R(\dots)$ and
$T(\dots)$ true, 40 have $R(\dots)$ true and $T(\dots)$ false, etc.
\end{example}




There is a separate counting context for each type. A \textbf{current context} is a set of pairs either of the form
$\tuple{Var,Value}$ where $Var$ is a PRV that has no parameters, or of
the form
$\tuple{\tau,CC}$ where $\tau$ is a type and $CC$ is a counting
context where the parameter of each of the PRVs in $CC$ is of type
$\tau$. A current context can have at most one pair for each type.

A PRV $P$ is \textbf{assigned} in a current context $Con$ if it has no
parameters and $\tuple{P,Val}\in Con$, or if it is parametrized by a
variable of type $\tau$ and if $\tuple{\tau,\tuple{V,\chi}}\in CC$
and $P$ unifies with an element of $V$.

A \textbf{parfactor graph} on $\tuple{Con,G}$ where $Con$
is a current context and $G$ is a set of parfactors, has the elements of $G$ as nodes, and there is an
arc between parfactors
$\tuple{C_1,V_1,\phi_1}$ and
$\tuple{C_2,V_2,\phi_2}$ if there is an element of $V_1$ that
isn't assigned in $Con$ that unifies with an element of $V_2$ such that the unifier does not violate any of
the constraints in either parfactor.

The \textbf{grounding} of a parfactor graph on $\tuple{Con,G}$ is a
factor graph on $\tuple{Con',G'}$, where for every counting parfactor
$\tuple{S,F,\theta}$ in $G$, and for every grounding substitution $\psi$
of all of the free parameters that does not violate $S$, $F\psi$ is in
$G'$ with table $\theta$.   $Con'$ represents all of the ground
instances that are assigned in $Con$, with the corresponding counts
given by the table in $Con$.

The grounding of a parfactor graph defines its semantics. We
carry out lifted operations so that the lifted operations have the same
result as carrying out recursive
conditioning on the grounding of the parfactor graph for the same
elimination ordering.

\subsection{Symmetry and Exchangeability}
The reason we can do lifted inference is because of symmetries. Having
a symmetry between the unnamed individuals means that a derivation
about some of the individuals can be equally applied to any of the
other individuals. 

We say that a set of individuals are \textbf{exchangeable} in a
parfactor graph if
the grounding of the parfactor graph with one consistent assignment of
individuals to variables is isomorphic to the grounding of the graph
with another assignment. Graph isomorphism means there is a 1-1 and onto
mapping between the nodes where the factors are identical. Exchangeability means that reasoning with some of the
individuals can be applied to the other individuals.


\subsection{Unification, splitting and shattering}
In order to determine which instances of parametrized random variables
are the same random variables,
\citet{reason:Poole03a} used unification and splitting on logical variables, which
guarantees that the instances are identical or are
disjoint. \Citet{reason:BraAmiRot05a} proposed to do all of the
splitting up-front in an operation called shattering
(see \citet{Kisynski:2009aa} for analysis of splitting, shattering and related operations).
\begin{figure}
\begin{center}
\includegraphics[scale=0.7]{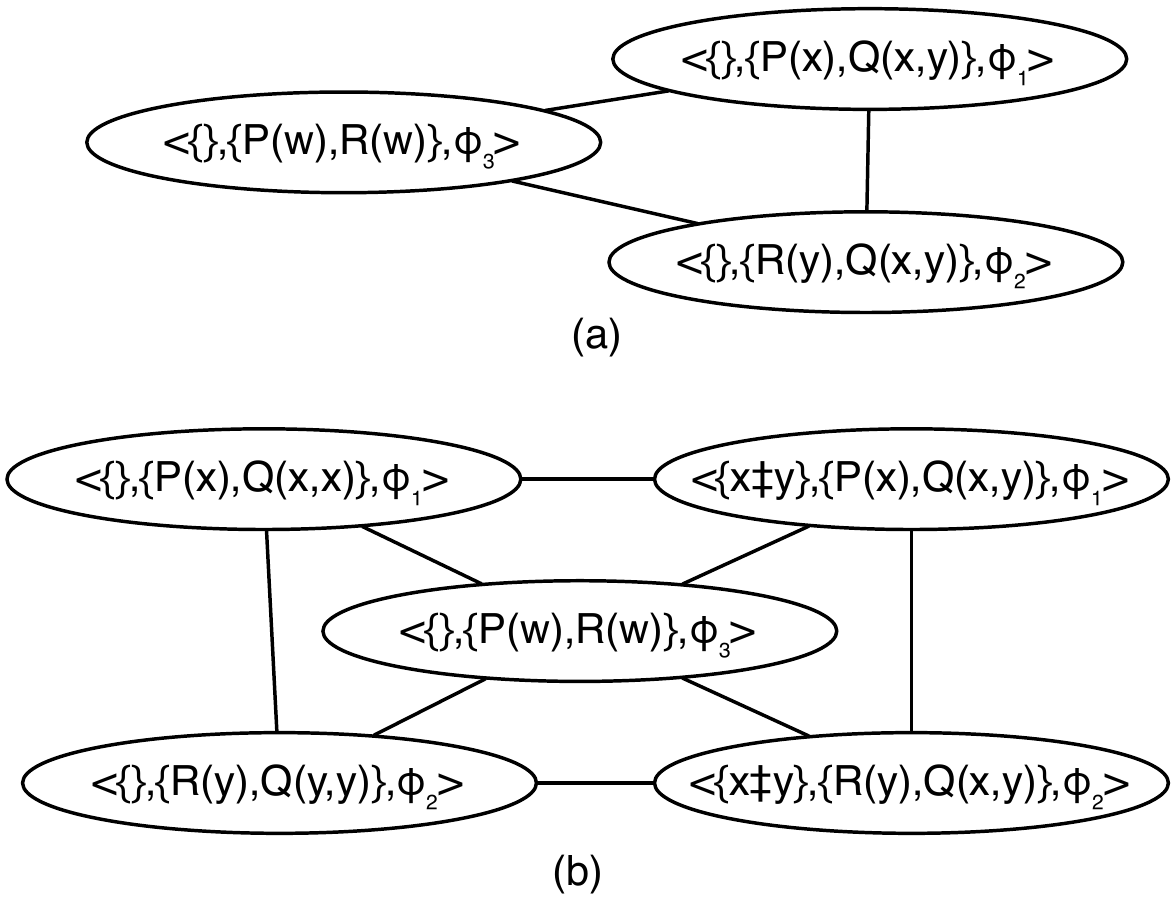}
\end{center}
\caption{A parfactor graph that is problematic for shattering and its
  preemptively shattered counterpart}
\label{shattering-counter-fig}
\end{figure}

Shattering is a local operation and does not imply
graphs constructed by substitutions are isomorphic as in the following
example:
\begin{example}
Consider the network of Figure
\ref{shattering-counter-fig} (a). Although it is shattered, the instances
with $x=y$ have a different grounding from the instances where
$x\neq y$. This graph can be split, on $x=y$ in the right-hand
parfactors, giving the network shown in Figure \ref{shattering-counter-fig} (b).
\end{example}

An alternative to the local shattering is to carry out a more global
preemptive shattering.
A set of parfactors \textbf{preemptive shattered} if 
\begin{itemize}
\item for every type,
and every constant $C$ of the type that is explicitly mentioned, every
parfactor that contains a variable $x$ of the type includes the
constraint $x\neq C$.
\item if variables $x$ and $y$ of the same type are in a parfactor,
  the parfactor contains the constraint $x\neq y$.
\end{itemize}
Given a set of parfactors, to construct an equivalent set of preemptively shattered set of parfactors, all
logical variables in a parfactor are split with respect to all
explicitly given constants, and any pairs of logical variables in a
parfactor are split with respect to each other.

Preemptive shattering gives more splits than shattering, and sometimes
more than needed, but it allows our proofs to work 
and does not prevent the asymptotic complexity results we
seek. With preemptive shattering, counting the number of instances
represented by a parfactor is straightforward; there are no complex interactions.  For
the rest of this paper, we assume that all parfactors are preemptively shattered.

Note that, as can be seen in the parfactor graph of Figure
\ref{shattering-counter-fig} (b), even after preemptive shattering, we cannot always globally rename
variables so that the unifying variables are identical. 

\subsection{Disconnected Grounding}
When the graph is disconnected, Algorithm \ref{RC-algo} considers the connected
components separately, and multiplies them. In this section, we cover
all of the cases where the grounding is disconnected, and show how
it corresponds to operations in the lifted case. 

If the lifted network is disconnected, the ground counterpart is
disconnected, and so these disconnected components can be solved
separately and multiplied.

If the lifted network is connected, this does not imply that its
grounding is connected. For example, the parfactor graph of Figure
\ref{ExampleRQ} (c) is connected yet its grounding is not connected.

Intuitively, if there is a logical variable that is in all of the
counting parfactors, 
the instances for one individual are disconnected  from the instances
for another individual. Thus, we can the solve the problem for one of the
individuals, and the value for the lifted case is that value to the
power of the number of individuals. 

This intuition needs to be refined because logical variables are local to a parfactor;
renaming the variables gives exactly the same grounding. There are
cases where chains of unifications cause connectedness:
\begin{example}The parfactor graph
\begin{center}
\includegraphics[scale=0.7]{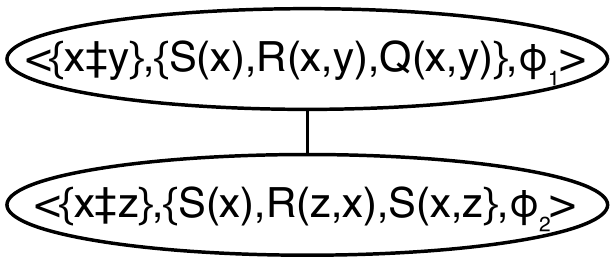}
\end{center}
does not have disconnected ground instances, even though $x$ is in
every PRV. For three individuals, $C_1, C_2, C_3$, in the grounding
$R(C_1,C_3)$ is connected to $R(C_2,C_3)$ through $S(C_3)$ in the
grounding of the bottom parfactor, thus
$S(C_1)$ is connected to $S(C_2)$ for any different $C_1$ and $C_2$,
using the top parfactor. 
\end{example}

This reasoning can be applied generally:

Suppose $x$ is a logical variable in parfactor $\tuple{C,V,\phi}$ that
appears in parfactor graph $G$.
$connected(x,\tuple{C,V,\phi},Con,G)$ means
the instances of $x$ in parfactor $\tuple{C,V,\phi}$ are connected to
each other in the grounding of $\tuple{Con,G}$. $connected$ can be
defined recursively as follows.

$connected(x,\tuple{C,V,\phi},Con,G)$ is true if and only if:
\begin{itemize}
\item $x$ appears in $V$, is not assigned in $Con$ and
 there is a PRV in $V$ that is not assigned in $Con$ and not parametrized by $x$ or
\item there is a parfactor $\tuple{C',V',\phi'}$ in $G$, such that an element of
  $V$ unifies with an element of $V'$ (in a manner consistent with $C$
  and $C'$, and that are not assigned in $Con$) and $x$ is unified with a
  variable $x'$ such that $connected(x',\tuple{C',V',\phi'},Con,G)$ is true.
\end{itemize}

The definition of $connected$ is sound: if $connected(x,\tuple{C,V,\phi},Con,G)$ is true,
the instances of $x$ are connected in the grounding.
The proof for the soundness is a straightforward induction proof; essentially
the algorithm is a constructive proof.

However, the definition is not complete: there can be instances that are connected
even though $connected$ is false. It is instructive
to see what a proof for completeness would look like. To prove
completeness, we would prove that all instances of $x$ are
disconnected if the above construction fails to derive they are
connected. Suppose there are two constants $C$ and $C'$, we need to
show that the graph with $x$ replaced by $C$ is disconnected with the
graph with $x$ replaced by $C'$. The graph with $x$ replaced by $C$
has $C$ in every PRV (by construction) and the graph with $x$ replaced
by $C'$ has $C'$ in every PRV. However, this does not imply that the
graphs are disconnected as there could be a PRV that contains both $C$
and $C'$, as in the following example:
\begin{example}\label{connected-eg}
Consider the parfactor graph:
\begin{center}
\includegraphics[scale=0.7]{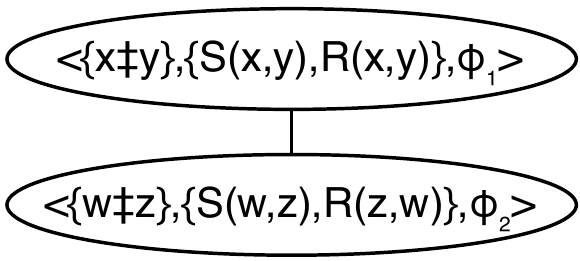}
\end{center}
In the grounding, for all individuals $C_i\neq C_j$, the random variable
$S(C_i,C_j)$ is connected to $S(C_j,C_i)$. However, it is disconnected from
other instances of $S(x,y)$.
\end{example}

We use the definition of $connected$ to detect potentially
disconnected components, and we can explicitly check for which
instances are connected. In this way, we can ensure that the lifted
algorithm detects
disconnectedness whenever the ground algorithm would. The only
counterexamples to the completeness of $connected$ are when there is a set of variables, all with the
same domain, and all of them appear in all PRVs in the parfactor graph
(possibly renamed), and there is an inequality constraint between
them. Suppose there are $k$ such variables, $x_1\dots,x_k$, in a parfactor. We choose
$k$ constants\footnote{These should be constants that don't otherwise
  appear in the current set of counting parfactors.
We need to choose the constants so that the
  same constants are used in different branches, to ensure that
  caching finds identical instances whenever the ground instances
  are found in the cache.}, $C_1,\dots,C_k$, and apply the substitution
$\{x_1/C_1,\dots,x_k/C_k\}$ to that parfactor, and then proceed to
ground out all the corresponding variables in the other factors by
unifying with the factors in all ways, forming a generic connected component. We then need to count the
number of copies of each PRV instance in the connected component;
suppose this is $c$. For a population size of $n$, there are $n(n-1)\dots(n-k+1)$
instances of the PRV, and there are $c$ elements in each connected
component, therefore there are $n(n-1)\dots(n-k+1)/c$ disconnected
components. If we compute
$p$ as the probability of the generic connected component, 
we need to take $p$ to the power $n!/((n-k)!\times c)$ to compute the
probability of the lifted network.

In the above analysis, $n$ is the population size and $k$ and $c$
depend only on the structure of the graph, and not on the population
size. As we assume that we can count the population size in time
logarithmic in the population size, the above procedure is polynomial
in the log of the population size, whereas grounding is polynomial in
the population size. As we see below, this is the only case where the
grounding is polynomial in the population size.

In Example
\ref{connected-eg}, $k=2$ and $c=2$, and so the power is $n(n-1)/2$. In an example with $k=3$, it is
possible that $c$ could be 1,2,3 or 6.

Algorithm \ref{LRC-algo3} gives the lifted variant of case 3 of
Algorithm \ref{RC-algo}. The main loop is the same as Algorithm
\ref{RC-algo}, with the recursive call $lrc(Con,Fs)$ where $Con$ is a
current context and $Fs$ is a set of
counting parfactors.

\begin{algorithm}
\DontPrintSemicolon
\SetAlgoNoEnd
\SetKwInput{Input}{input}
\SetKwInput{Output}{output}
\SetKwFunction{proc}{def}
\SetKwBlock{Begin}{}{}
\If(\{\emph{Case 3a}\})
{$Fs$ is disconnected}{
     $\Return \prod_{\mbox{connected component }cc}lrc(Con,cc)$\;}
\ElseIf(\{\emph{Case 3b}\}){$\exists x~\neg connected(x,F,Con,Fs)$ for $F\in Fs
  $}{
Select one $x$ such that $\neg connected(x,F,Con,Fs)$\;
let $\tau$ be the type of $x$\;
let $C=\{ x \mbox{ of type }\tau:\neg connected(x,F,Con,Fs)\} \mbox{ for some }F\in Fs$\;
let $k=|C|$\;
replace $x_1,\dots,x_k$ with $C_1,\dots,C_k$ in $F$\;
unify $F$ with all other factors in $Fs$\;
let $c=|\{$instances of $F$ in $Fs\}|$\;
let $e=n!/((n-k)!\times c)$\;
\Return{$lrc(Con,Fs)^{e}$}\;
}
\caption{Lifted search, case 3: grounding is disconnected.}
\label{LRC-algo3}
\end{algorithm}

\subsection{Counting}
Once we have a single connected component (and there is no logical
variable for which its instances are connected), we select a PRV to
branch on. In describing this, as in Algorithm \ref{RC-algo}, we decouple
branching on a variable, and evaluating parfactors. Typically a PRV
is associated with many parfactors, and when we branch on a PRV we need to count the number of instances with various values for the
PRV. We need to make sure that we branch in a way that enables us to
evaluate the relavant parfactors.

For the simplest case, assume we want to sum out a Boolean PRV $F(x)$ that has
one free logical variable, $x$, with domain $\{A_1,\dots,A_n\}$. The
idea behind counting \citep{de-Salvo-Braz:2007aa} is that only the
number of exchangeable individuals that have a PRV having a particular
value matters, not
their identity. We present counting first considering this simple
case, then
more complex cases.

In \textbf{counting branching}, for each $i$, such that $0\leq i \leq
n$, the algorithm generate the branch where there are $i$ instances of $F$ true, and so
$n-i$ instances of $F$ false.
This branch represents 
$\binom{n}{i}$ paths in the grounding, as there are this many renamings
of constants that would result in the same assignment. Thus it can
multiply the result of evaluating this branch by $\binom{n}{i}$.
Note that the counting branching involves generating
$n+1$ branches, whereas in the grounding 
there are $2^n$ 
assignments of values after the equivalent ground branching.
The resulting counting context records the number of
instances of $F$ that are true and number that are false.

We now show how to evaluate various cases of parfactors that can
include $F$. The general case is a combination of these specific
cases. For all of the example below, assume they are part of a larger
parametrized graphical model. In particular, assume that the instances are
connected, so that the case described in the previous section does not apply.

\begin{example}\label{FxE-ex}
Consider the parfactor
\[\tuple{\{\},\{F(x),E\},\phi_1}\]
Suppose $|x|=n$. This parfactor represents $n$ factors. Suppose $\phi_1$ is:
\[\begin{array}{ll|l}
F(x)&E& \phi_1\\\hline
True & True & \alpha_1\\
True & False & \alpha_2\\
False & True & \alpha_3\\
False & False & \alpha_4
\end{array}\]
Suppose we have split on $E$ and assigned it the value $True$,and then
we split on $F(x)$ and are in the branch with $F=True$ for $i$ cases and
$F=False$ for $n-i$ cases. This is represented by the current context:
\[E=true, \begin{array}{l|l}
F(x)& \phi_1\\\hline
True  & i\\
False  & n-i\\
\end{array}\]
The contribution of this parfactor in this current context is
$\alpha_1^i\alpha_3^{n-i}$.
\end{example}

\begin{example}\label{xydifftypeseg}
Consider the parfactor
\[\tuple{\{\},\{F(x),G(y)\},\phi_2}\]
Suppose $x$ and $y$ are of different types, where $|x|=n$,
$|y|=m$. $\phi_2$ is:
\[\begin{array}{ll|l}
F(x)&G(y)& \phi_2\\\hline
True & True & \alpha_1\\
True & False & \alpha_2\\
False & True & \alpha_3\\
False & False & \alpha_4
\end{array}\]
This
parfactor represents $nm$ factors; for each combination of assignments
of values to the instances of $F$ and $G$, there is a factor. In a
current context with 
$i$ $F$'s true and $h$ $G$'s true, this parfactor has a contribution: 
\[\alpha_1^{ih}\alpha_2^{i(m-h)}\alpha_3^{(n-i)h}\alpha_4^{(n-i)(m-h)}.\]
\end{example}

\begin{example}\label{FxGx-ex}
Consider the parfactor
\[\tuple{\{\},\{F(x),G(x)\},\phi_2}\]
Suppose $|x|=n$. This parfactor represents $n$ factors. Unlike the
previous cases, counting branching is not adequate; we need to consider
which $F$-assignments go with which
$G$-assignments. We can do a counting branch on $F$
first: for each $i \in [0,n]$, consider the case where $F$ is true for
$i$ individuals, and is false for $n-i$ individuals. This case represents
$\binom{n}{i}$ branches in the grounding. To split on $G$ we can do a
dependent branch: consider the $i$ individuals for which $F$ is true, and the $n-i$ individuals
for which $F$ is false separately. For each
$j\in [0,i]$,  we consider the branch where $G$ is true for $j$
individuals and  false for $i-j$ individuals all with $F=True$; this branch
corresponds to $\binom{i}{j}$ ground branches. For each $k\in
[0,n-i]$ we construct the branch where $G$ is true for $k$ individuals
and is false for $i-j-k$ individuals with $F=False$. This represents
$\binom{n-i}{k}$ ground cases. This branch is represented by
the counting context:
\[\begin{array}{ll|l}
F(x)&G(x)& \phi_2\\\hline
True & True & j\\
True & False & i-j\\
False & True & k\\
False & False & n-i-k
\end{array}\]
The contribution of the parfactor in this branch is:
\[\alpha_1^{j}\alpha_2^{i-j}\alpha_3^{k}\alpha_4^{n-i-k}\]

\end{example}

\begin{example}\label{xneyFGex}
Consider the parfactor
\[\tuple{\{x\neq y\},\{F(x),G(y)\},\phi_2}\]
Suppose $x$ and $y$ are of the same type, where $|x|=|y|=n$. This
parfactor represents $n(n-1)$ factors. This can be solved by a mix of
the previous two examples.
If we were to do the same as Example \ref{xydifftypeseg}, we would
also include the cases where $x=y$, which are explicitly excluded; but these are 
the cases in Example \ref{FxGx-ex}. So the contribution of this factor can be computed by
dividing the result of Example \ref{xydifftypeseg} by the result of
Example \ref{FxGx-ex}, or equivalently subtracting the exponents. As in Example \ref{FxGx-ex}, we consider the
case where $F$ is true for $i$ individuals, and for these individuals $G$ is
true for $j$ of them, and out of the individuals where $F$ is false, $G$ is
true for $k$ of them. Taking the difference between the exponents
Example \ref{xydifftypeseg} and  \ref{FxGx-ex}, and noticing that $h$
in Example  \ref{xydifftypeseg} corresponds to $j+k$ in Example
\ref{FxGx-ex}, the contribution of these factors is:
\[\alpha_1^{i(j+k)-j}\alpha_2^{i(n-j-k)-i+j}\alpha_3^{(n-i)(j+k)-k}\alpha_4^{(n-i)(n-j-k)-n+i+k}\]
\end{example}


\begin{example}\label{fghex}
Consider a mix between the previous examples. Suppose we have the parfactors:
\[\tuple{\{\dots\},\{F(x),G(x),\dots\}}\]
\[\tuple{\{x\neq y,\dots\},\{H(y),G(x),\dots\}}\]
where all of the variables are of
  the same type with $n$ individuals. Suppose the branching order is to branch on
  $H$, then $F$, then $G$. The split on $G$ needs to depend on both $H$ and
  $F$. This can be done if the splits on $H$ and $F$ are dependent;
  that is, we do a separate count on $F$ for the individuals for which
  $H$ are true and the individuals for which $H$ are false. Then we do
  a separate count\footnote{\label{lifted-con-footnote}Note that if we
  had projected the counts onto the separate factors, we would have
  lost the interdependence between $F$ and $H$, which is needed as the
count for $G$ depends on both.} on $G$ for the set of individuals for each
  combination of values to $H$ and $F$.
\end{example}

Counting branching needs to be expanded to cascaded counting branching. 
\textbf{Dependent counting branching} on a PRV $X$ that is parametrized by a
parameter of type $\tau$, works as follows. First,
we find the corresponding counting context
$\tuple{V,\chi}$ for $\tau$. Dependent Counting branching replaces
this with a
counting context on $\tuple{V\cup \{X\},\chi'}$ as follows. For each
assignment $\tuple{t,i}$ in the table $\chi$ ($i$ is the count
for assignment $t$), for each $j$
in $[0,i]$, we create the table $\chi'$ that maps $t\cup\{X=true\}$ to $i$ and
$t\cup\{X=false\}$ to $i-j$. This assignment corresponds to $\binom{i}{j}$,
different grounding assignments, so the grounding needs to be
multiplied by $\binom{i}{j}$. This is recursively carried out for each tuple.

\begin{example}
Starting from the current context of Example \ref{FxE-ex}, dependent counting
branching on $G(x)$ produces the counting context of Example \ref{FxGx-ex}.
This context corresponds to $\binom{i}{j}\binom{n-i}{k}$ contexts in
the grounding. Note that there are $i(n-i)$ leaves that are decedents
of the current context created in Example \ref{FxE-ex}, whereas in the
grounding there are $2^n$ leaves that are descendants of each
corresponding ground context.
\end{example}

Branching is shown as Case 4 in Algorithm
\ref{Lifted-RC-algo}. In this algorithm $Con$ is a current context and
$Fs$ is a set of input parfactors. Case 4a is the same as case 4 in
Algorithm \ref{RC-algo} (but for Boolean variables). Case 4b is for
branching on a PRV with a single parameter, and sets up
dependent counting branching that is presented in Algorithm
\ref{branch-algo}. Note that this treats a counting context as a set
of pairs of an assignment of values to a set of PRVs and a count (as
described in Section \ref{IntRepSec}). 

The
branching factor depends on the population, but the depth of the recursive calls depends on the
structure of the counting context, and not on the population size. The
depth of the recursive calls provides the power of the polynomial. If
we use a sparse representation of current contexts with zeros suppressed, this is never worse
than grounding. [However, whether we use a sparse or dense
representation is something that can be optimized for.]

\begin{algorithm}
\DontPrintSemicolon
\SetAlgoNoEnd
\SetKwInput{Input}{input}
\SetKwInput{Output}{output}
\SetKwFunction{proc}{def}
\SetKwBlock{Begin}{}{}
\Input{$Con$: current context\;
$Fs$: set of parfactors
}
\Output{a number representing $\sum_{\overline{x}} \prod_{F \in grounding(Fs,Con)}
  F(\overline{x})$}
\If(\{\emph{Case 0}\})
{$\exists x \in vars(Con) \setminus vars(Fs)$}{
\Return{$lrc(\sum_X Con,Fs)$}
}
\If(\{\emph{Case 1}\})
{$\exists v$ such that $\tuple{\tuple{Con,Fs},v}\in cache$}{
\Return{$v$}}
\ElseIf(\{\emph{Case 2}\}){$\exists f\in Fs: vars(f)\subseteq vars(Con)$}{
     \Return{ $ eval\_parfactor(f,Con)\times lrc(Con,Fs\setminus\{f\})$}\;
}
\Else(\{\emph{Case 3}\}){See Algorithm \ref{LRC-algo3}}{
}
\Else(\{\emph{Case 4}\}){
select PRV $X \in vars(Fs)\setminus vars(Con)$\;
\If(\{\emph{Case 4a}\}){$X$ contains no parameters}{
$sum \gets lrc(\{X=true\}\cup Con,Fs) + lrc(\{X=false\}\cup Con,Fs)$\;
}
\Else(\{\emph{Case 4b}\}){
suppose the parameter of $X$ is of type $\tau$\;
\If{$\exists \chi:~\tuple{\tau,\chi} \in Con$}
{
$sum \gets  branch(\chi,X,\{\},Con\setminus \{\tuple{\tau,\chi}\},Fs)$\;
}
\Else{
$sum\gets  branch(\{\tuple{\tuple{},|\tau|}\},X,\{\},Con,Fs)$\;
}
}
$cache \gets cache \cup \{\tuple{\tuple{Con,Fs},sum}\}$\;
\Return{$sum$}\;
}
\caption{Lifted Recursive Conditioning: $lrc(Con,Fs)$}
\label{Lifted-RC-algo}
\end{algorithm}

\begin{algorithm}
\DontPrintSemicolon
\SetAlgoNoEnd
\SetKwInput{Input}{input}
\SetKwInput{Output}{output}
\SetKwFunction{proc}{def}
\SetKwBlock{Begin}{}{}
\Input{$\chi$: a set of tuples from a counting context\;
$X$: the PRV to branch on its instances\;
$\chi'$: the new counting context being constructed\;
$Con$: the current context to be added to\;
$Fs$: the set of all factors\;
}
\If
{$\chi=\{\}$}{
Suppose $\tau$ is the type of the parameter in $V$\;
\Return{$lrc(Con\cup\{\tuple{\tau,\chi'}\},Fs)$}\;
}
\Else{
select $\tuple{t,i}\in \chi$\;
$sum \gets 0$\;
\For{$j \in [0,i]$}{
let $\chi''$ be $\chi'\cup\tuple{t\cup\{X=true\},j}\cup\tuple{t\cup\{X=false\},i-j}$\;
$sum \gets sum+\binom{i}{j} branch(\chi\setminus \{\tuple{t,i}\},\chi'',Con,Fs)$\;
}
\Return{sum}\;
}
\caption{Dependent Counting Branching:\\$branch(\chi,X,\chi',Con,Fs)$}
\label{branch-algo}
\end{algorithm}

The main remaining part of the lifted algorithm is to  evaluate a parfactor $\tuple{C,V,\phi}$ in a counting context $\tuple{V',\chi}$, where all of
the variables in $V$ are assigned in $V'$. There are three cases: shared
parameters, different parameters of the same type and parameters of
different types. One parfactor can contain all of these.

For shared parameters, as in Example \ref{FxGx-ex}, the
parfactor
provides the base, and there is a unique counting context that
provides the powers. First we group all of these together and raise
them to the appropriate powers, and then treat them as a block.

For parameters of different types, as in Example \ref{xydifftypeseg},
we need to multiply the powers. We can treat the shared parameters as
a block.

For different parameters of the same type, as in Example
\ref{xneyFGex}, we can use the other two cases: first we treat them as
different types (which over-counts because it includes the equality
cases), and then divide by the case when they are equal. We also have
to readjust for double counting, which can be done using the
coefficient of $n!/(n-k)!$ where $k$ is the number of such cases. For
example, when $k=3$, this is $n(n-1)(n-2)=n^3-3n^2+2n$. The first of
these corresponds to all parameters being different, the second to all pairs
of parameters equal, and the third to all parameter the same. 

Algorithm \ref{evalpf-algo} shows how to evaluate a parfactor in a
current context. It omits the last case, as it is computed from the other two cases.

\begin{algorithm}
\DontPrintSemicolon
\SetAlgoNoEnd
\SetKwInput{Input}{input}
\SetKwInput{Output}{output}
\SetKwFunction{proc}{def}
\SetKwBlock{Begin}{}{}
\Input{$PF$: a parfactor\;
$Con$: a current context\;
}
Suppose $PF$ is $\tuple{C,V,\phi}$\;
Suppose $Con$ is $\tuple{V',\chi}$\;
$prod \gets 1$\;
\ForEach
{$\tuple{t,p}\in \phi$}{
\If{$t$ is consistent with variable assignments in $Con$}{
$power\gets 1$\;
Let $\tau$ be the type of $X$\;
Select $\tuple{\tau,\tuple{V'',\chi}}\in Con$\;
$RedundantVars \gets vars(V'') \setminus vars(V)$\;
$power \gets power \times \sum_{\tuple{t,i}\in \chi: consistent(t,V)}i$\;
}
$prod \gets prod \times p^{power}$
}
\Return{$prod$}\;
\caption{Evaluating a parfactor in current context:\\$eval\_parfactor(PF,Con)$}
\label{evalpf-algo}
\end{algorithm}

\paragraph{Example \ref{fghex} (cont.)}
Consider the branch where $H$ is true for $i$ individuals and false
  for $n-i$ individuals. Suppose we then branch on $F$. We then consider the branch with 
    $F$ true for $j_0$ of the cases where $H$ is false and $j_1$ cases
    where $H$ is true. We thus have: $j_1$ individuals for which $F$
    and $H$ are true; $i-j_1$ individuals which have $H$ true and $F$
    false; $j_0$ individuals that have $H$ false and $F$ true; and
    $n-i-j_0$ individuals what have both $H$ and $F$ false. 
We can then branch on $G$, for each of the four
    sets of individuals. We thus know the counts of each case;
    Algorithm \ref{evalpf-algo} 
    computes the contribution of each factor.



The final two cases of the algorithm are caching (case 1 of Algorithm
\ref{RC-algo}) and forgetting (case 0 of Algorithm
\ref{RC-algo}). Caching can remain the same, we just have to ensure
that the cache can find elements that are the same up to renaming of
variables, which can be done easily as the current context does not
depend on the name and the variable and can be stored in a canonical
way (e.g., alphabetically). Forgetting is the inverse of splitting. A
variable in a counting context that doesn't appear in the parfactors
can be summed out of the counting context (which is the same operation
as summing out a variable in variable elimination). $\sum_X Con$ in Algorithm \ref{Lifted-RC-algo}
means to sum out $X$ from the counting context it appears in or to
remove it if it is not a parametrized variable.

This description assumed binary-valued variables, and only functors with
0 or 1 arguments. The first of these is straightforward to generalize, and the second is not.

Consider what happens when $F$ can have more than
two values. Suppose $F$ is a unary $m$-valued PRV with range
$\{v_1,\dots,v_m\}$. That is, $F(A_i)$ is a random variable with
domain $\{v_1,\dots,v_m\}$. The assignments we need to consider are
when there are non-negative integers $i_1\dots i_m$ where $i_m$
represents the number of individuals that have value $i$. Thus for
each assignment to $i_1\dots i_m$, where $i_j\geq 0$ for each $j$ and
$i_1+\dots+i_m=n$, we consider the assignment 
\begin{align*}
F(A_i)=v_{1} &\mbox{ for } 0<i\leq i_1\\
F(A_i)=v_{2} &\mbox{ for } i_1<i\leq i_1+i_2\\
&\dots\\
F(A_i)=v_{m} &\mbox{ for } i_1+i_2+\dots + i_{m-1}<i\leq n
\end{align*}
It is a straightforward combinatorial exercise to include this in the
algorithm (but complicates the description).


\section{Conclusion}

Lifted probabilistic reasoning has proved to be challenging. There
have been many proposals to lift various algorithms, however all of the
exact algorithms needed to ground out a population in some
cases (and it is often difficult to tell for which cases they need to
ground a population). We set out to determine if there was some fundamental
reason why we would need to ground out the representation, or whether
there was some case where we needed to effectively ground out. We
believe that we have answered this for two cases:
\begin{itemize}
\item When lifted inference is polynomial in a population, which
  occurs when VE does not create a factor that is parametrized by a
  population or search can be disconnected for a population, we can
  solve it in time polynomial in the logarithm of the population.
\item For parametrized random variables with zero or a single
  argument, and search-based inference (and so also variable
  elimination, due to their equivalent complexity) is
  exponential when grounding, we answer arbitrary conditional queries
  in time polynomial in the population.
\end{itemize}

The question of whether we can \emph{always} do lifted inference in polynomial time
in each population size when there are PRVs with more than one argument, is still an open
problem. While we can use the algorithm in this paper for many of
these cases, there are some very tricky cases. Hopefully the results
in this paper will provide tools to fully solve this problem.

We have chosen to not give empirical comparisons of our results. These
are much more comparisons of the low-level engineering than of the
lifted algorithm. There are no published algorithms that can correctly
solve all of the examples in this paper in a fully lifted form.


\end{document}